\documentclass[letter]{svproc}
\usepackage{url}

\usepackage{times}
\usepackage{fixltx2e}
\usepackage[numbers]{natbib}
\usepackage{amsmath}
\usepackage{mathabx}
\usepackage{multicol}
\usepackage[bookmarks=true]{hyperref}
\usepackage{graphicx}
\usepackage[utf8]{inputenc}
\usepackage{graphics} % for pdf, bitmapped graphics files
\usepackage{epsfig} % for postscript graphics files
\usepackage{mathptmx} % assumes new font selection scheme installed
\usepackage{bm}
\usepackage{caption}
\usepackage{subcaption}
\usepackage{ifthen}
\newboolean{include-notes}
\usepackage[linesnumbered,ruled,noend]{algorithm2e}
\usepackage[noend]{algpseudocode}
\usepackage{hyperref}
\usepackage{pst-node,pst-plot}

\DeclareMathOperator*{\argmin}{arg\,min}

\begin{document}
\mainmatter              % start of a contribution
\title{Pushing Fast and Slow: Task-Adaptive Planning for Non-prehensile Manipulation Under Uncertainty}
\titlerunning{Pushing Fast and Slow}  % abbreviated title (for running head)
%                                     also used for the TOC unless
%                                     \toctitle is used
%

\author{Wisdom C. Agboh \hspace {2mm} \and \hspace {2mm} Mehmet R. Dogar }
\authorrunning{WC Agboh and MR Dogar} % abbreviated author list (for running head)
%
%%%% list of authors for the TOC (use if author list has to be modified)
\tocauthor{Wisdom C. Agboh, Mehmet R. Dogar}
\institute{School of Computing, University of Leeds \\ Leeds, LS29JT, United Kingdom,\\
\email{\{scwca,m.r.dogar\}@leeds.ac.uk}
}

\maketitle              % typeset the title of the contribution

\begin{abstract}
We propose a planning and control approach to physics-based 
manipulation. The key feature of the algorithm is that it can adapt to the
accuracy requirements of a task, by slowing down and generating ``careful''
motion when the task requires high accuracy, and by speeding up and moving fast
when the task tolerates inaccuracy. We formulate the problem as an MDP with action-dependent stochasticity and propose an
approximate online solution to it. We use a trajectory optimizer with a
deterministic model to suggest promising actions to the MDP, to reduce
computation time spent on evaluating different actions. We conducted experiments in simulation and on a real robotic system. Our results show that with a task-adaptive planning and control approach, a 
robot can choose fast or slow actions depending on the task accuracy and uncertainty level. The robot makes these decisions online and is able to maintain high success rates while completing manipulation tasks as fast as possible. 
\end{abstract}
\section{INTRODUCTION}
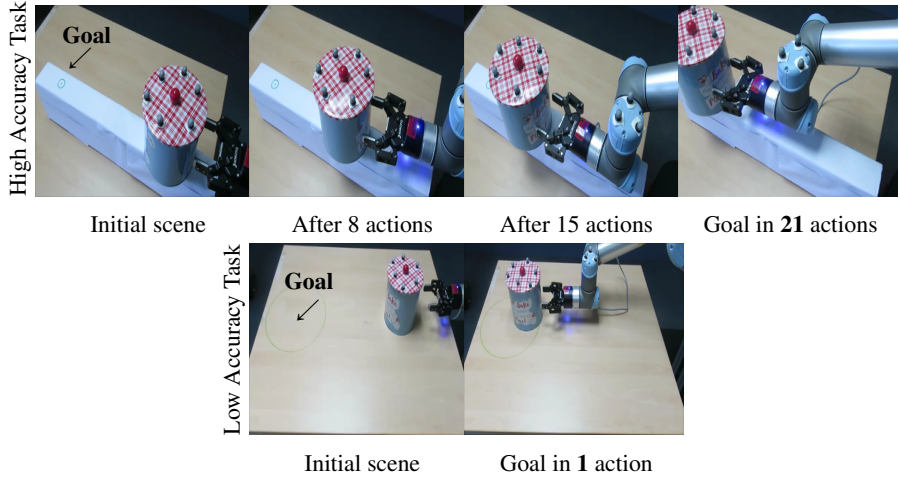
\begin{figure}[htb!]
	\centering
    \begin{subfigure}[b]{0.245\textwidth}
   \includegraphics[height=1.0in,width=1.15in]{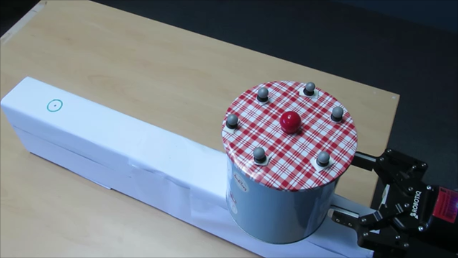}
   \begin{picture}(0,0)
      \put(-95.0,0.0){\rotatebox{90}{High Accuracy Task}}
      \put(-75,50){  $\swarrow$ }
      \put(-75,58){\textbf{Goal}}
    \end{picture}
    \caption*{Initial scene}
    \label{fig:high_accuracy_task}
  \end{subfigure}
  \hspace{-3mm}
  \begin{subfigure}[b]{0.245\textwidth}
    \includegraphics[height=1.0in,width=1.15in]{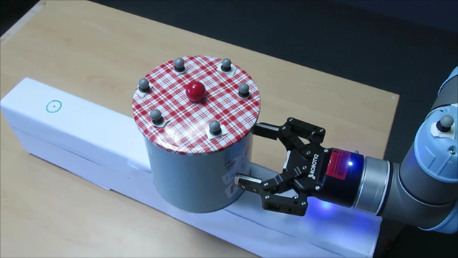}
    \caption*{After 8 actions}
  \end{subfigure}
  \hspace{-3mm}
  \begin{subfigure}[b]{0.245\textwidth}
    \includegraphics[height=1.0in,width=1.15in]{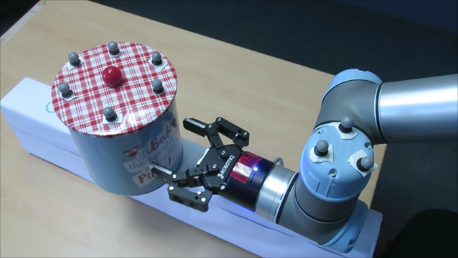}
    \caption*{After 15 actions}
  \end{subfigure}
  \hspace{-3mm}
  \begin{subfigure}[b]{0.245\textwidth}
    \includegraphics[height=1.0in,width=1.15in]{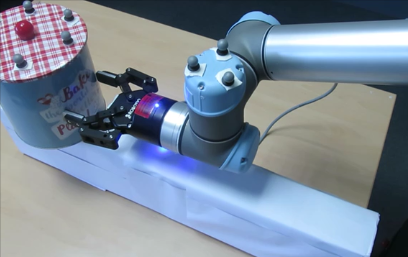}
    \caption*{Goal in \textbf{21} actions}
  \end{subfigure} 

\begin{subfigure}[b]{0.245\textwidth}
    \includegraphics[height=1.0in,width=1.15in]{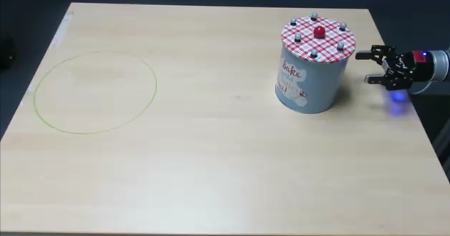}
       \begin{picture}(0,0)
      \put(-95.0,0.0){\rotatebox{90}{Low Accuracy Task}}
       \put(-70,45){  $\swarrow$ }
      \put(-72,55){\textbf{Goal}}
    \end{picture}
    \caption*{Initial scene}
    \label{fig:low_accuracy_task}
  \end{subfigure}
  \hspace{-3mm}
  \begin{subfigure}[b]{0.245\textwidth}
    \includegraphics[height=1.0in,width=1.15in]{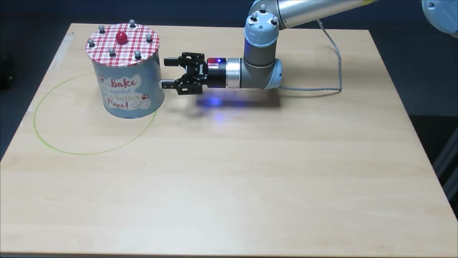}
    \caption*{Goal in \textbf{1} action}
  \end{subfigure} 
  \caption{Task-adaptive pushing with 21 slow actions for a high accuracy task (top) and a single fast action for a low accuracy task (bottom). }\label{fig:fig1}
   \vspace{-6mm}
 \end{figure}

\vspace{-2mm}
We propose a planning and control algorithm for non-prehensile
manipulation.  The key feature of our algorithm is \textit{task-adaptivity}:
the planner can adapt to the accuracy requirements of a task, performing
fast or slow pushes. For example in Fig.~\ref{fig:fig1} (top), the robot is
pushing an object on a narrow strip. The task requires high-accuracy during
pushing --- otherwise the object can fall down. The controller therefore
generates slow pushing actions that make small but careful progress to the goal
pose of the object. In Fig.~\ref{fig:fig1} (bottom), however, the object is on a
wide table and the goal region for the object is large (circle drawn on the
table). In this case, the controller generates only a small number of fast
pushing actions to reach the goal quickly --- even if this creates more
uncertainty about the object's pose after each action, the task can still be
completed successfully. We present a controller that can adapt to tasks with
different accuracy requirements, such as in these examples.
There has been significant recent interest in non-prehensile pushing-based
manipulation. 
Most existing work use motion planning and \textit{open-loop} execution to address
the problem of generating a sequence of actions to complete a non-prehensile
manipulation task\citep{king2015nonprehensile, kitaev_abbeel,
dogar_clutter,randomized_clutter_uncertainty,convergent_planning}. 
Others developed closed-loop approaches. \citet{effect_aware_push} proposed a compact predictive model for sliding objects and a feedback
controller to push an object to a desired location.
\citet{pusher_slider} proposed a model predictive controller that uses integer programming to handle the different contact modes associated with  the pusher-slider system. \citet{mppi_push} considered the task of pushing
an object to a goal region while avoiding regions of high predictive
uncertainty through model predictive path integral control. We take a similar model predictive control (MPC) approach and propose a closed-loop planning and control algorithm for pushing tasks. 

A common feature of existing work is the reliance on the  quasi-static model of
pushing \citep{mason_pushing,howe1996practical}. While one reason of the
popularity of the quasi-static model may be the simpler analytic equations of
motion it enables one to derive, another reason is the slow nature of
quasi-static interactions, which keeps the uncertainty during pushing tightly
bounded and therefore easier to control accurately.

However, accuracy is not the main criterion for every task, as we illustrate in
Fig.~\ref{fig:fig1}. Fast motions, even if inaccurate, may be desired during
some tasks. 
We humans also adapt our actions to the task (Fitts's law \cite{fitts_law}). Imagine reaching into a fridge shelf that is crowded with fragile objects, such as glass jars and containers.
You move slowly and carefully. However, if the fridge shelf is almost empty,
only with a few  plastic containers that are difficult-to-break, you move faster
with less care. 

The major requirements to build a task-adaptive planner/controller are: 
\vspace{-2mm}
\begin{enumerate}
\item The planner must \textbf{consider a variety of actions (different pushing speeds)}:
The robot should not be limited to moving at quasi-static speeds. It must consider dynamic actions wherever possible to complete a given task as fast as possible. 
  
\item The planner must \textbf{consider action-dependent uncertainty}: Different actions can induce different amounts of uncertainty into the system. For example, pushing an object for a longer distance (or equivalently pushing faster for a fixed amount of time) would induce more uncertainty than pushing a short distance (or equivalently pushing slower for a fixed amount of time) \cite{million_ways_to_push}.

\end{enumerate}

One way to build such a controller is to model the problem as a Markov Decision
Process (MDP) with stochastic dynamics, where the stochasticity is action-dependent.
Then, if this MDP is solved for an optimal policy under a
cost that includes time to reach the goal, the resulting policy will use fast
actions when it can, and fall back to slow actions for tasks that require higher
accuracy.

In this paper, we model the problem as an MDP. However, we do not search for a
globally optimal policy as this would be prohibitively computationally
expensive. Instead, we solve the MDP online \citep{peret2004line, kearns} with an approximate solution. Even in this online setting, evaluating the value of all possible actions
(including actions of a wide variety of speeds), proves computationally
expensive, since the cost of physics-based predictions is high.
Therefore, instead of evaluating all possible actions, at any given state, we
first use a fast trajectory optimizer to suggest a reduced set of promising
actions, i.e. actions that are known to drive the system to the goal under the
deterministic setting. We then 
evaluate these actions under uncertainty to pick the best one. 

Our specific contributions include a task-adaptive online solution to the stochastic MDP for pushing-based manipulation and a trajectory optimizer to generate actions for evaluation under the MDP setting. Additionally, we compare our task-adaptive planner with a standard model predictive control approach using only slow actions for high and low accuracy tasks under different levels of uncertainty. We show that our approach achieves higher success rates in significantly smaller amounts of time, especially for tasks that do not require high accuracy. Finally, we implement our approach on a real robotic system for tasks requiring different accuracy levels and compare it with standard MPC. Results can be found in the video at 
\url{https://youtu.be/8rz_f_V0WJA}.
\section{PROBLEM FORMULATION}
\label{problemdescription}

We consider the problem where a robot must plan a sequence of non-prehensile
actions to take an environment from an initial configuration to a desired goal
configuration. We consider two task categories: In the \textit{pushing} task
the goal is to push a target object into a goal region; and in the \textit{grasping
in clutter} task the goal is to bring a target object, among other objects, into the robot's hand.
Our scenes contain $D$ dynamic objects. $\vec{q}^{i}$ refers to the full pose of each dynamic object, for $i=1, \dots, D$. We assume a flat working
surface and the robot is not allowed to drop objects off the
edges.

Our robot is planar with a 1-DOF gripper. The robot's configuration is defined by a vector of joint values ${\vec{q}^{R} = \{\theta_{x}, \theta_{y}, \theta_{rotation}, \theta_{gripper}\}}$. We represent the complete state of our system as $\vec{x}_{t}$ at time $t$. This includes the configuration and velocity of the robot and all dynamic objects; $\vec{x}_{t} = \{\vec{q}^{R}, \vec{q}^{1}, \dots, \vec{q}^{D},\dot{\vec{q}}^{R},\dot{\vec{q}}^{1}, \dots, \dot{\vec{q}}^{D}\}$. Our control inputs are velocities: $\vec{u}_{t}=\dot{\vec{q}}^{R}$ applied to the robot's joints. We then define the stochastic discrete time dynamics of our system as:
\begin{equation}\label{eq:dynamics}
 \vec{x}_{{t+1}} = f(\vec{x}_{{t}}, \vec{u}_{t}) + \zeta(\vec{u}_{t})
\end{equation}
where $f$  is a deterministic function that describes the evolution of state  $\vec{x}_{t}$ given the action $\vec{u}_{t}$.  We induce stochasticity in the system dynamics through $\zeta(\vec{u}_{t}) \propto ||\vec{u}_{t}||$, which is proportional to the magnitude of action $\vec{u}_{t}$.  When we push an object over a long distance, there is a large number of interactions/contacts especially in cluttered environments. This implies that the uncertainty in the resulting state at the end of a long push should be larger than that for a shorter push. 

We assume an initial state of the system $\vec{x}_{0}$. Our goal is to generate a sequence of actions for the robot such that the desired final goal configuration of the environment is reached as quickly as possible without dropping objects off the edge of our working surface. In the foregoing paragraphs, $\vec{U}=\{\vec{u}_{0},\vec{u}_{1}, \dots, \vec{u}_{n-1}\}$ denotes a sequence of control signals of fixed duration $\Delta_{t}$ applied in $n$ time steps and we use brackets to refer to the control at a certain time step, i.e. $\vec{U}[t] =\vec{u}_{t} $. Similarly, $\vec{X}$ is a sequence of states. 

To build a task-adaptive controller, we formulate our problem as an MDP, and we
provide an approximate solution to it.  
An MDP is defined by a tuple $<S,A,P,L'>$, where $S$ is the set of states, $A$
is the set of actions, $P$ is the probabilistic transition function, and $L'$ 
defines the costs. In our problem $S$ is given by all possible values of
$\vec{x}_{t}$. Similarly, $A$  is given by all
possible values of $\vec{u}_{t}$, and $P$ can be computed using the stochastic transition function in Eq.~\ref{eq:dynamics}.
\noindent The optimal policy for an MDP is given by:
\vspace{-2mm}
\begin{align}
\pi^*(\vec{x}_{t}) = \argmin_{\vec{u}_{t} \in A} \left[ L'(\vec{x}_{t},\vec{u}_{t}) + \gamma \cdot \int_{S} P(\vec{x}_{t+1}|\vec{x}_{t},\vec{u}_{t}) \cdot V^*(\vec{x}_{t+1}) \cdot dx_{t+1}\right]
%\pi^*(\vec{x}_{t}) = \argmin_{\vec{u}_{t} \in A} \left[ L(\vec{x}_{t},\vec{x}_{t+1},\vec{u}_{t}) + \gamma \cdot \sum_{\vec{x}_{t+1} \in S} P(\vec{x}_{t+1}|\vec{x}_{t},\vec{u}_{t}) \cdot V^*(\vec{x}_{t+1}) \right]
\end{align}
where $0 < \gamma < 1$ is the discount factor, and $V^*$ is the optimal value function. 
\noindent An online one-step lookahead approximate solution to the MDP problem can be found by sampling
and evaluating the average value over samples as
in \cite{peret2004line,kearns}:
\vspace{-2mm}
\begin{align}
\widetilde{\pi}(\vec{x}_{t}) = \argmin_{\vec{u}_{t} \in A} \left[ L'(\vec{x}_{t}, \vec{u}_{t}) + \frac{1}{Q} \cdot \sum_{\vec{x}_{t+1} \in S(\vec{x}_{t},\vec{u}_{t},Q)} \widetilde{V}(\vec{x}_{t+1}) \right]
\end{align} 
where $S(\vec{x}_{t},\vec{u}_{t},Q)$ is the set of $Q$ samples found by stochastically propagating
$(\vec{x}_{t},\vec{u}_{t})$, and $\widetilde{V}$ is an approximation of the value function. For our problem, to compute the cost $L'(\vec{x}_{t}, \vec{u}_{t})$,
we use a cost function $L$ which also takes into account the next state
$\vec{x}_{{t+1}}$:
\vspace{-2mm}
\begin{align}\label{eq:cost}
L(\vec{x}_{{t}}, \vec{x}_{{t+1}}, \vec{u}_{{t}}) = \sum_i^D \{w_{e} \cdot e^{k\cdot d^{i}_{P}} + w_{s} \cdot (\vec{x}_{t+1}^{i} - \vec{x}_{t}^{i})^2 \} + k_{act}
\end{align}
\noindent The first term in the cost which we call the \textit{edge cost} penalizes pushing an object close to the table's boundaries or static obstacles. We show the edge cost in Fig.~\ref{fig:edgecost} where we define a safe zone smaller than the table's boundaries. If an object is pushed out of this safe region as a result of an action between $t$ and $t+1$, we compute the pushed distance $d_p$. Also note that $k$ is a constant term and no edge costs are computed for objects in the safe zone. 
The second term is the \textit{environment disturbance cost} which penalizes moving dynamic objects away from their current states.
The third term, $k_{act}$ is a constant cost incurred for each action taken by the robot. We use $w_e$ and $w_s$ to represent weights for the edge and environment disturbance costs respectively. 
Then, we compute $L'(\vec{x}_{t}, \vec{u}_{t})$ using the same set of Q samples:
\vspace{-2mm}
$$
L'(\vec{x}_{t}, \vec{u}_{t}) =  \frac{1}{Q} \cdot \sum_{\vec{x}_{t+1} \in S(\vec{x}_{t},\vec{u}_{t},Q)} L(\vec{x}_{{t}}, \vec{x}_{{t+1}}, \vec{u}_{{t}})
$$
This solution requires propagating $Q$ samples for every possible action $\vec{u_{t}}$, to find the one with the minimum total cost. Performing this for all actions $\vec{u}_{t} \in A$ is not feasible for our purposes for a variety of reasons: First, we are interested in actions that span a wide range of speed profiles (i.e. fast
and slow), which make our potential action set large; second, each propagation in our domain is a physics simulation which is computationally expensive; and third, our goal is closed-loop pushing behaviour close to real-time speeds.
%%%%%%%%%%%%%%%%%%%%%%%%%%%%%%%%%%%%%%%%%%%%%%%%%%%%%%%%%%%%%%%
\section{PROPOSED APPROACH}
\vspace{-2mm}
\label{sec:mdp}
%%%%%%%%%%%%%%%%%%%%%%%%%%%%%%%%%%%%%%%%%%%%%%%%%
\begin{figure}[t]
	\vspace{6mm}
	\centering 
    \includegraphics[height=2.2in,width=4.8in]{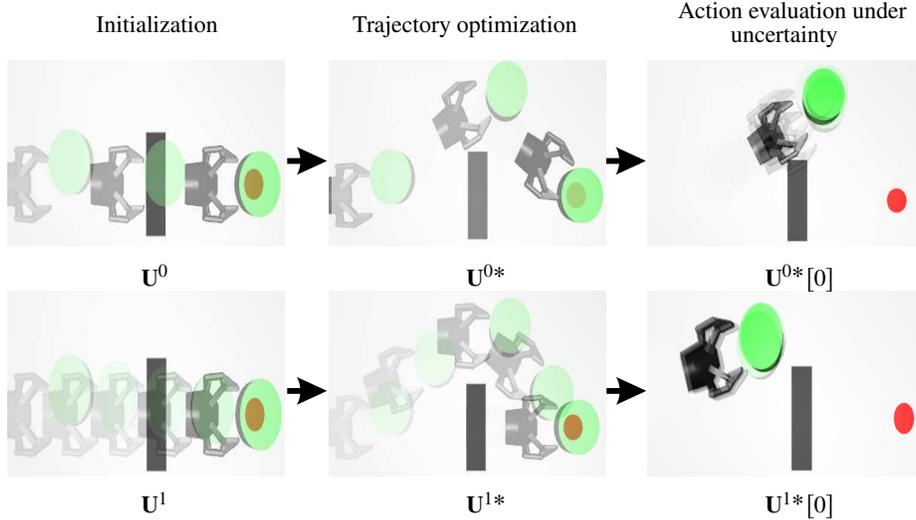}
    \begin{picture}(0,0)
      \put(-140,180){Initialization}
	  \put(-43,180){Trajectory optimization}
	  \put(80,185){Action evaluation under}
	  \put(100,176){uncertainty}
	  
	  %\put(-123,165){$\vec{u}^{0}$}
	  \put(-123,84){$\vec{U}^{0}$}
	  \put(-123,-2){$\vec{U}^{1}$}
	  
	  %\put(0,165){$\vec{u}^{0*}$}
	  \put(0,84){$\vec{U}^{0*}$}
	  \put(0,-2){$\vec{U}^{1*}$}
	  
	  %\put(123,165){$\vec{u}^{0*}_{0}$}
	  \put(113,84){$\vec{U}^{0*}[0]$}
	  \put(113,-2){$\vec{U}^{1*}[0]$}
	  
    \end{picture}
  \caption{First column: Initialization of our task-adaptive approach with control sequences including fast (top) and slow (bottom) actions. Second column: Stochastic trajectory optimization of the initial candidate control sequences. Last column: Action evaluation under uncertainty through sampling. }\label{fig:summary}
  \vspace{-4mm}
 \end{figure}
%%%%%%%%%%%%%%%%%%%%%%%%%%%%%%%%%%%%%%%%%%%%%%%%%%%%%%%%%%%%
Instead of considering a large action set in our online MDP solution (Eq. 3), we propose to use
a small set of promising actions including both fast and slow actions. We identify such a
set of actions using a trajectory optimizer based on the deterministic dynamics function
$f$ of the system. 

Note that our approach does not discretize the action space or the state space \textit{a priori}. We adaptively sample the action space using a trajectory optimizer to find high value actions to consider at a given current state. We also get stochastic next state samples by applying these actions through a physics simulator.

In Alg.~1, we present our online approximate solution to the MDP.

Consider the scene in Fig.~\ref{fig:summary}, where we have a planar gripper and an object. Our task is to push the object to a desired goal location (the red spot) while avoiding the rectangular black obstacle. We begin by generating $N$ candidate action sequences $\{\vec{U}^{0}, \dots, \vec{U}^{N-1}\}$ to the goal by using the $GetActionSequences$  procedure (line 1). The number of actions in each sequence varies between $n_{min}$ and $n_{max}$, and each action is of fixed duration $\Delta_{t}$. In the example task (Fig.~\ref{fig:summary}), we show the candidate action sequences in the first column where $N=2$, $n_{min}=2$, and $n_{max}=4$. Since each action is of fixed duration, the set of action sequences contain both fast (top) and slow (bottom) actions. Details of how we generate these candidate control sequences are explained in Sec~\ref{sec:generating_actions}. 

Using a trajectory optimizer (Sec.~\ref{sec:trajectory_optimization}), the procedure $GetOptActionSequences$ returns $N$ optimized control sequences $\{\vec{U}^{0*}, \cdots , \vec{U}^{N-1*}\}$, and an approximation of the value function $\{\widetilde{\vec{V}}^{0}, \cdots,\widetilde{\vec{V}}^{N-1}\} $ along the optimal trajectories. We  visualize the optimized trajectories for the example task in the second column. 

\setlength{\textfloatsep}{2mm}
\begin{algorithm}[t]\label{alg:mdpa}
    \SetKwInOut{Input}{Input}
    \SetKwInOut{Output}{Output}
    \SetKwInOut{Parameters}{Parameters}
    \SetKwInOut{Subroutines}{Subroutines}

    \Input{
     $\vec{x}_{0}$: Initial state}
    \Parameters{
     $Q$: Number of stochastic samples\\
     $N$: Number of initial candidate control sequences\\ 
     $n_{min}$: Minimum number of control actions in a control sequence\\
	 $n_{max}$: Maximum number of control actions in a control sequence 
     }
    $\{\vec{U}^{0}, \dots, \vec{U}^{N-1}\} \gets  $GetActionSequences$(\vec{x}_{0},n_{min}, n_{max}, N)$ \\
	\While {task not complete} { 	
	
     $\{\vec{U}^{0*}, \cdots , \vec{U}^{N-1*}\} , \hspace{1mm} \{\widetilde{\vec{V}}^{0}, \cdots,\widetilde{\vec{V}}^{N-1}\} \gets $GetOptActionSequences$(\vec{x}_{0},\{\vec{U}^{0}, \dots, \vec{U}^{N-1}\})$ \\
   
     \For{$i \gets 0$ \KwTo $N-1$, }
     {
     $V^{i}=0$

     \For{each sample in $Q$, }
     {

     $\vec{x}_{1}= $StochasticExecution$(\vec{x}_{0},\vec{U}^{i*}[0]) $

     $V^{i} = V^{i} + L(\vec{x}_{0},\vec{x}_{1},\vec{U}^{i*}[0]) + \widetilde{\vec{V}}^{i}[0]$

     }
     $V^i = V^i / Q$

     }
     $i_{min}$ = $\argmin_{i \in N} V^{i} $

     $\vec{x}_{1} \gets$ execute $\vec{U}^{i_{min}}[0]$
     
     check task completion \\
     $\vec{x}_{0} \gets \vec{x}_{1}$
    
     $\{\vec{U}^{0}, \dots, \vec{U}^{N-1}\} \gets  \{$GetActionSequences$(n_{min}, n_{max}, N-1),\hspace{1mm}
         \vec{U}^{i_{min}}[1:n-1]\}$   
     
     }

    \caption{Online MDP solver}
\end{algorithm}
\setlength{\floatsep}{2mm}
We seek a one-step lookahead solution to the $MDP$, hence, we get the first actions $\vec{U}^{i*}[0]$ from each of the optimized control sequences. Our task is now to select the best amongst $N$ actions even under uncertainty. We apply each action $Q$ times to the stochastic state transition function in Eq.~\ref{eq:dynamics} to yield $Q$ next state samples (line 7). More specifically, using our system dynamics model, we apply the controls (velocities) to the robot for the control duration $\Delta_{t}$ and thereafter we wait for a fixed extra time $t_{rest}$ for objects to come to rest before returning the next state and computing the cost. We can see the $Q$ samples in the third column (Fig.~\ref{fig:summary}) for the example pushing task. Thereafter, on line 8, for each sample we add the immediate cost $L$ to the approximate value $\widetilde{\vec{V}}^{i}[0]$ for the resulting state. We use the same approximate value for all $Q$ samples. We then compute an average value for each action (line 9), select the best one and execute it (line 11).  If the task is not yet complete, we repeat the whole process but also re-use the remaining portion of the best control sequence $ \vec{U}^{i_{min}}[1:n-1]$ from the current iteration. 
This algorithm chooses slow, low velocity actions for tasks that require high accuracy but faster actions for tasks that allow inaccuracies. 

In the example pushing task, Fig.~\ref{fig:summary} (third column), we see a wider distribution in the resulting state for the fast action (top) compared to the slower action (bottom) as a result of our uncertainty model. Such a wide distribution in the resulting state increases the probability of undesired events happening especially in high accuracy tasks. For example, we see that some samples for the fast action result in collisions between the robot and the obstacle. This implies high costs with respect to the slower action, hence our planner chooses the slow action in this case, executes it and starts the whole process again from the resulting state. 
\vspace{-3mm}
\section{GENERATING A VARIERTY OF ACTIONS}
\vspace{-2mm}
\label{sec:generating_actions}
\setlength{\textfloatsep}{2mm}
\begin{algorithm}[t]
    \SetKwInOut{Input}{Input}
    \SetKwInOut{Output}{Output}
    \SetKwInOut{Parameters}{Parameters}
    \SetKwInOut{Subroutines}{Subroutines}
    \Output{$\{\vec{U}^{0}, \dots, \vec{U}^{N-1}\}$: A set of candidate action sequences}
    \Parameters{$\Delta_{t}$: Control duration for each action in an action sequence
    }
 
     \For{$k \gets 0 $ \KwTo $N-1$ }
     {
     $n^{k} = \lceil{ \dfrac{n_{max} - n_{min}}{N-1}k}\rceil + n_{min}$ \\
		
     $\vec{U}^{k}[0:n^{k}-1] = \{\dfrac{Distance\hspace{1mm}to\hspace{1mm}goal}{n^{k}\Delta_{t}}\} \vec{1}$

     }

     \Return $\{\vec{U}^{0}, \dots, \vec{U}^{N-1}\}$

    \caption{GetActionSequences ($\vec{x}_{0},n_{min},n_{max},N$)}
    \label{alg:get_action_sequences}
\end{algorithm}
\setlength{\floatsep}{2mm}
\setlength{\textfloatsep}{2mm}
\begin{algorithm}[t]
    \SetKwInOut{Input}{Input}
    \SetKwInOut{Output}{Output}
    \SetKwInOut{Parameters}{Parameters}
    \SetKwInOut{Subroutines}{Subroutines}
    \Output{$\{\vec{U}^{0*}, \cdots , \vec{U}^{N-1*}\}$: Optimized set of action sequences \\
  	$\{\widetilde{\vec{V}}^{0}, \dots, \widetilde{\vec{V}}^{N-1} \} $ \Comment{ Approx. value function along N optimal trajectories }}
     \For{$i \gets 0 $ \KwTo $N-1$ }
     {

     $\vec{U}^{i*}, \widetilde{\vec{V}}^{i} \gets $TrajectoryOptimization($\vec{x}_{0}, \vec{U}^{i}$)

     }

     \Return $\{\vec{U}^{0*}, \cdots , \vec{U}^{N-1*}\} , \hspace{1mm} \{\widetilde{\vec{V}}^{0}, \cdots,\widetilde{\vec{V}}^{N-1}\}$

    \caption{GetOptActionSequences($ \vec{x}_{0},\{\vec{U}^{0}, \dots, \vec{U}^{N-1}\}$)}\label{alg:get_opt_sequences}
\end{algorithm}
\setlength{\floatsep}{2mm}
At each iteration of our online MDP solver, we provide $N$ actions that are evaluated under uncertainty. First, using the $GetActionSequences$ procedure in Alg.~\ref{alg:get_action_sequences}, we generate $N$ candidate action sequences where the number of actions in an action sequence increases linearly from $n_{min}$ to $n_{max}$ (line 2).

 On line 3, each of the action sequences is set to a straight line constant velocity profile to the goal. This is a simple approach, other more complicated velocity profiles could also be used here. Furthermore, in the $GetOptActionSequences$ procedure, Alg.~\ref{alg:get_opt_sequences}, we use these candidate action sequences to initialize a stochastic trajectory optimization algorithm (Sec.~\ref{sec:trajectory_optimization}). The algorithm quickly finds and returns a locally optimal solution for each of the candidate control sequences. It also returns an approximation of the value function along $N$ optimal trajectories. We use this approximate value while evaluating actions in our online MDP solver. 
\section{TRAJECTORY OPTIMIZATION}
\vspace{-2mm}
\label{sec:trajectory_optimization}
Trajectory optimization involves finding an optimal control sequence $\vec{U}^{*}$ for a planning horizon $n$, given an initial state $\vec{x}_{0}$, an initial candidate control sequence $\vec{U}$, and an objective $J$ which can be written as: 
\vspace{-2mm}
\begin{align}
\label{eq:cost_function}
J(\vec{X},\vec{U}) = \sum_{t=0}^{n-1} L(\vec{x}_{t},\vec{x}_{t+1},\vec{u}_{t}) + w_{f}L_{f}(\vec{x}_{n})
\end{align}
$J$ is obtained by applying the control sequence $\vec{U}$ starting from a given initial state and includes the sum of running costs $L$ and a final cost $L_{f}$. We use the constant $w_{f}$ to weight the terminal cost with respect to the running cost. We consider a deterministic environment defined by the state transition function $\vec{x}_{t+1} = f(\vec{x}_{t}, \vec{u}_t)$. This is a constraint that must be satisfied at all times.
Then the output of trajectory optimization is the minimizing control sequence: 
\vspace{-2mm}
\begin{align}
    \vec{U}^{*} =arg \min \limits_{\vec{U}} J(\vec{X},\vec{U})
\end{align}

In this work, we consider stochastic trajectory optimization methods such as STOMP \citep{kalakrishnan2011stomp} and MPPI \citep{mppi}. With parallel rollouts on multiple cores of a PC, these methods show impressive speed. They also easily accept arbitrary cost functions that do not have to be differentiable. In contrast with sampling-based methods such as RRTs and PRMs \citep{king2015nonprehensile,randomized_clutter_uncertainty}, optimization approaches are able to produce lower cost trajectories within a time limit even if they do not take the system to the desired goal state. These benefits make stochastic trajectory optimization very attractive. In this work, we propose Alg.~\ref{alg:sto} which adapts the STOMP algorithm \citep{kalakrishnan2011stomp} for non-prehensile object manipulation. 
\setlength{\textfloatsep}{2mm}
\begin{algorithm}[t]
    \SetKwInOut{Input}{Input}
    \SetKwInOut{Output}{Output}
    \SetKwInOut{Parameters}{Parameters}
    \SetKwInOut{Subroutines}{Subroutines}

    \Input{$\vec{x}_{0}$: Initial state \\ 
     $\vec{U}$: Candidate control sequence containing $n$ actions }
    \Output{$\vec{U}^{*}$: Optimal control sequence}
    \Parameters{
     $K$: Number of noisy trajectory rollouts \\
     $\vec{\nu}$: Sampling variance vector \\
     $C_{thresh}$: Success definition in terms of cost\\
      $I_{max}$: Maximum number of iterations}
   	 $\vec{X} , \vec{C} \gets $TrajectoryRollout$(\vec{x}_{0},\vec{U})$ \\
     \While{$I_{max}$ not reached \textbf{and} Sum$(\vec{C}) > C_{thresh}$}{
     \For{$k \gets 0 $ \KwTo $K-1$ }
     {
     $\vec{\delta{U}}^{k} \gets N(\vec{0},\vec{\nu})$  \Comment{Random control sequence variation}  \\
     $\vec{U}^{k} = \vec{U} + \vec{\delta{U}}^{k}$ \\ 
     $\vec{X}^{k} , \vec{C}^{k} \gets $TrajectoryRollout$(\vec{x}_{0},\vec{U}^{k})$\\
      }
 	$\vec{U}^{*} \gets $UpdateTrajectory$(\vec{U},\{\vec{\delta{U}}^{0}, \dots , \vec{\delta{U}}^{K-1} \}, \{\vec{C}^{0}, \dots \vec{C}^{K-1}   \})$ \\
 	
 	$\vec{X}^{*}, \vec{C}^{*} \gets $TrajectoryRollout$(\vec{x}_{0} , \vec{U}^{*})$
    
     \If {Sum$(\vec{C}^{*}) <  $ Sum$(\vec{C})$} 
     {
         $\vec{U} \gets \vec{U}^{*}$, \hspace{3mm} $\vec{X} \gets \vec{X}^{*} $, \hspace{3mm} $\vec{C} \gets \vec{C}^{*}$ \\
     }
     
 }

\For{$j \gets 0$ \KwTo $n-1$ }
{
 $\widetilde{\vec{V}}[j] = \sum_{h=j}^{n-1} \vec{C}^{*}[h]$ \Comment{ Approx. value function for state $\vec{x}^{*}_{j}$ along the optimal trajectory }
}

\Return $\vec{U}^{*} , \widetilde{\vec{V}}$
 
    \caption{Stochastic Trajectory Optimization}\label{alg:sto}
\end{algorithm}
\setlength{\floatsep}{2mm}
\setlength{\textfloatsep}{2mm}
\begin{algorithm}[ht!]
	\SetKwInOut{Input}{Input}
	\SetKwInOut{Output}{Output}
	\SetKwInOut{Parameters}{Parameters}
	\SetKwInOut{Subroutines}{Subroutines}
	\Input{$ \vec{x}_{0}$: Initial state \\ 
		$\vec{U}$: Control sequence with $n$ actions }
	
	\Output{$\vec{X}: $State sequence \\
		$\vec{C}: $Costs along the trajectory}
	
	\For{$t \gets 0 $ \KwTo $n-1$ }
	{
	
	$\vec{x}_{t+1} = f(\vec{x}_{t} , \vec{u}_{t})$ \\
	$\vec{C}[t] =L(\vec{x}_{t}, \vec{x}_{t+1}, \vec{u}_{t})$ \Comment{Calculate cost using Eq.~\ref{eq:cost}} \\ 
	
	\If{$t == n-1$}
{
	$\vec{C}[t] = \vec{C}[t] + L_{f}(\vec{x}_{t+1})$ \Comment{Add final cost}

}
		
	}
	
	\Return $\vec{X}, \vec{C}$
	
	\caption{TrajectoryRolllout}\label{alg:trajectory_rollout}
\end{algorithm}
\setlength{\floatsep}{2mm}

We begin with an initial candidate control sequence $\vec{U}$ and iteratively seek lower cost trajectories (lines 2-10) until the cost reaches a threshold or until the maximum number of iterations is reached (line 2). We add random control sequence variations $\vec{\delta U}^{k}$ on the candidate control sequence to generate $K$ new control sequences at each iteration (line 5). Thereafter on line 6, we do a trajectory rollout for each sample control sequence using the $TrajectoryRollout$ procedure in~Alg.~\ref{alg:trajectory_rollout}. It returns the corresponding state sequence $X$ and costs $C$ calculated for each state along the resulting trajectory. I.e $C[t]$ is the cost of applying action $u_{t}$ in state $x_{t}$.  After generating $K$ sample control sequences and their corresponding costs, the next step is to update the candidate control sequence using the $UpdateTrajectory$ procedure. One way to do this is a straightforward greedy approach where the minimum cost trajectory is selected as the update:
\vspace{-2mm}
\begin{align}
k{*} =arg \min \limits_{k} \sum_{t=0}^{n-1} \vec{C}^{k}[t] \hspace{3mm}, \hspace{10mm}
\vec{U}^{*} = \vec{U} + \vec{\delta U}^{k*}
\end{align}
\vspace{-2mm}
\noindent Another approach is a cost-weighted convex combination similar to \citep{mppi}: 
\begin{align}
\vec{U}^{*}[t] = \vec{U}[t] + \dfrac{\sum_{k=0}^{K-1} [\exp(-(\frac{1}{\lambda})\vec{C}^{k}[t])]\vec{\delta U}^{k}[t] }{\sum_{k=0}^{K-1} \exp(-(\frac{1}{\lambda})\vec{C}^{k}[t])}
\end{align}
\noindent Where $\lambda$ is a parameter to regulate the exponentiated cost's sensitivity. In our experiments, for a small number of noisy trajectory rollouts $K$ (e.g. $K=8$), a greedy update performs better. Hence, we use the greedy update in all our experiments.  

Once the trajectory update step is complete, we update the candidate control sequence only if the new sequence has a lower cost (line 9, Alg. ~\ref{alg:sto}). The trajectory optimization algorithm then returns the locally optimal control sequence and an approximation of the value function for each state along the trajectory, where the value function is approximated using the sum of costs starting from that state.  

\begin{figure}[t]
\centering
\begin{minipage}{.5\columnwidth}
  \centering
  \includegraphics[width=.4\linewidth]{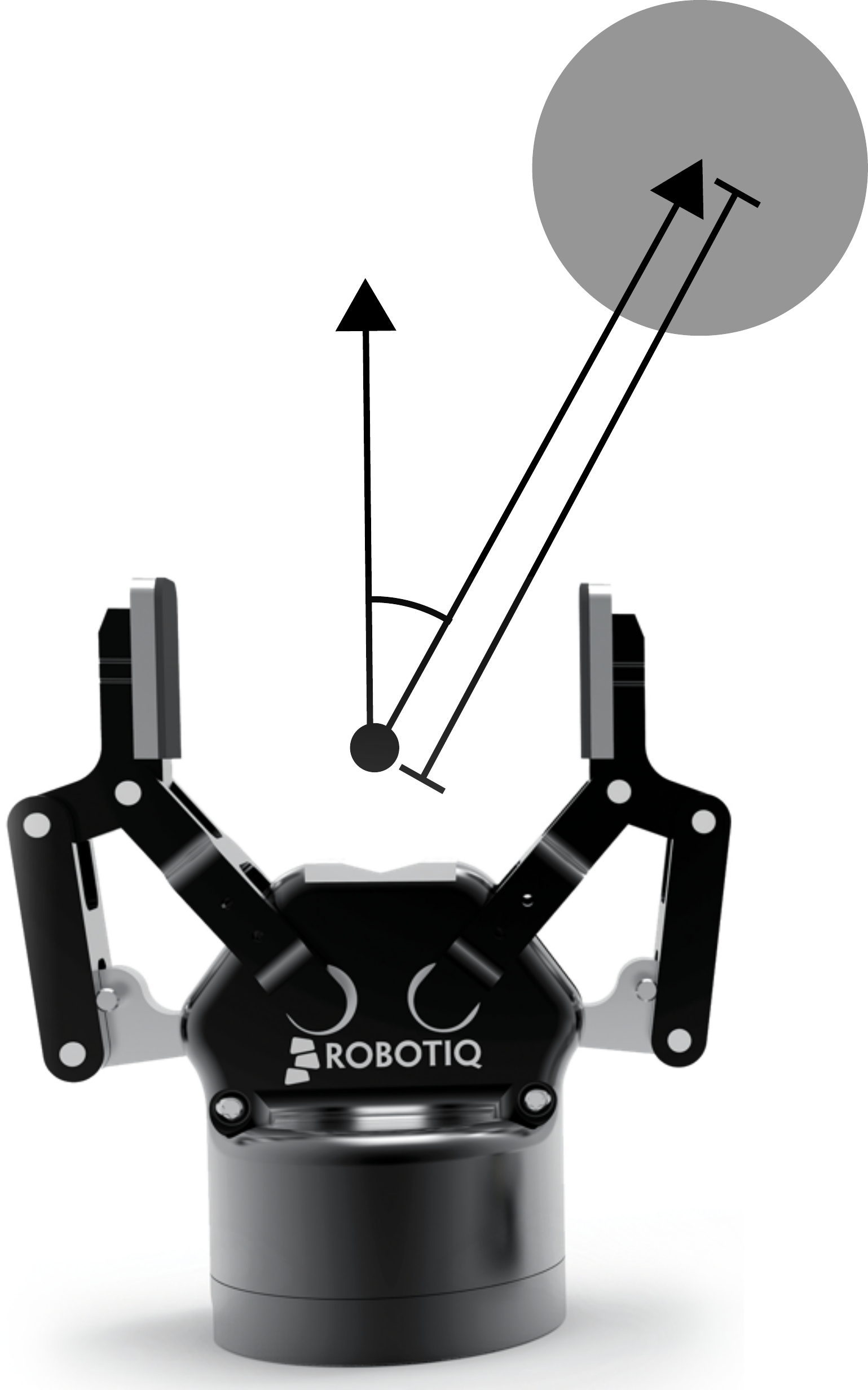}
     \begin{picture}(0,0)
      \put(-55.0,95.0){Target}
      \put(-40.0,71.0){\small{$\phi_T$}}
      \put(-17,73){$d_T$}
    \end{picture}
  \captionof{figure}{Goal cost terms}
  \label{fig:goalcost}
\end{minipage}%
\begin{minipage}{.5\columnwidth}
  \centering
  \includegraphics[width=.63\linewidth]{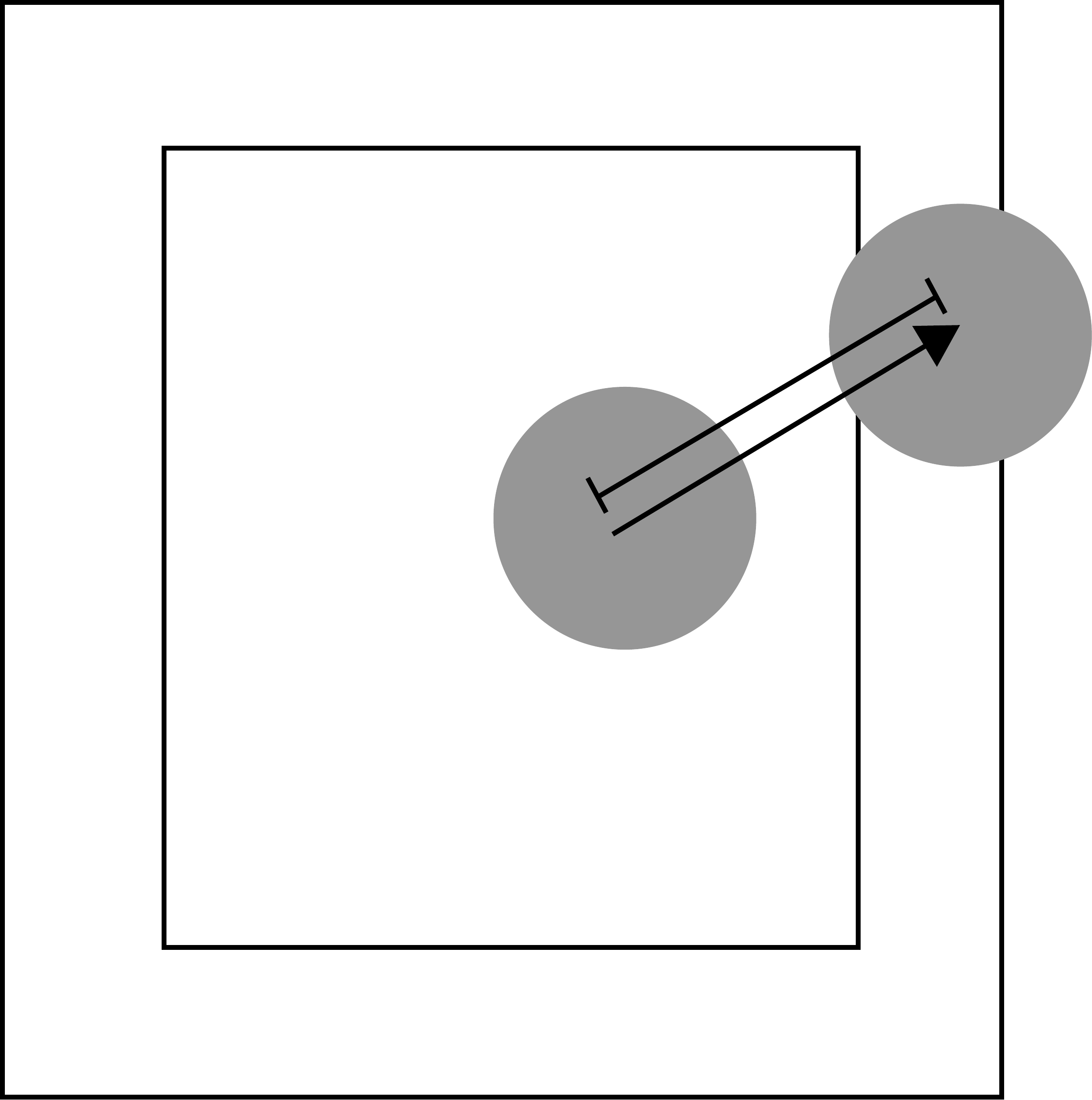}
     \begin{picture}(0,0)
      \put(-80.0,25.0){Safe zone}
      \put(-70.0,3.0){Table}
      \put(-40.0,75.0){$d_P^i$}
      \put(-48.0,37.0){$\vec{x}_{t}^i$}
      \put(-8.0,54.0){$\vec{x}_{t+1}^i$}
    \end{picture}
  \captionof{figure}{Edge cost terms}
  \label{fig:edgecost}
\end{minipage}
\end{figure}

The cost terms for the state-action sequences in this algorithm are equal to the running costs in Eq.~\ref{eq:cost}, with the addition of a terminal cost on the final state depending on the task. 
The terminal cost for the pushing task is given by:
\vspace{-2mm}
 \[
  L_{f} =
  \begin{cases}
    0 & \text{if $R_{o} - R_{g} < 0 $} \\
    (R_{o} - R_{g})^2 & \text{if $R_{o} - R_{g} > 0$}
  \end{cases}
\]
where $R_o$ is the distance between the pushed object and
the center of a circular goal region of radius $R_g$. The terminal cost term for the task of grasping in clutter is given by:
$L_{f}=d_{T}^2+w_{\phi}\cdot \phi_{T}^2$. We show how the distance $d_{T}$ and the angle $\phi_{T}$ are computed in Fig.~\ref{fig:goalcost}. First, create a vector from a fixed point in the gripper to the target object where $d_T$ is the length of this vector and $\phi_{T}$ is the angle between the forward direction of the gripper and the vector. We use $w_{\phi}$ to weight angles relative to distances. 
\section{BASELINE APPROACH}
%\vspace{-2mm}
\label{sec:baseline}
We implement a standard model predictive control algorithm ($MPC$) as a baseline approach. It involves repeatedly solving a finite horizon optimal control problem using the stochastic trajectory optimizer presented in Alg.~\ref{alg:sto} and proceeds as follows: optimize a finite horizon control sequence, execute the first action, get the resulting state of the environment and then re-optimize to find a new control sequence. When re-optimizing, we warm-start the optimization with the remaining portion of the control sequence from the previous iteration such that optimization now becomes faster. We initialize the trajectory optimizer for standard MPC with actions at the quasi-static speed. In addition, we propose another baseline approach to compare against in this work: uncertainty aware model predictive control (UAMPC). This is a version of our online MDP solver where only low speed actions are considered. Specifically, all the candidate control sequences are generated with the maximum number of actions i.e. $n_{min} \gets n_{max}$. 
%%%%%%%%%%%%%%%%%%%%%%%%%%%%%%%%%%%%%%%%%%%%%%%%%%%%%%%%%%%%%%%%%
\section{EXPERIMENTS}
\vspace{-2mm}
\label{experiments}
%%%%%%%%%%%%%%%%%%%%%%%%%%%%%%%%%%%%%%%%%%%%%%%%%%%%%%%%%%%%%%%%%%%%

We call our planning approach task-adaptive model predictive control ($TAMPC$). We verify how well our approach is able to handle uncertainty and adapt to varying tasks. First, we compare the performance of $TAMPC$ with a standard model predictive control ($MPC$) approach. Here we hypothesize that $TAMPC$ will complete a given pushing task within a significantly shorter period of time and will be able to adapt to different tasks, maintaining a high success rate under varying levels of uncertainty. 

Next, we compare the performance of our approach with uncertainty aware MPC ($UAMPC$). Here we hypothesize that: $UAMPC$ will have a similar success rate and will take a longer amount of time to complete the task in comparison with $TAMPC$. 

%Furthermore, we investigate whether using our algorithm, a robot can exhibit a truly task-adaptive behaviour in different environments: executing fast dynamic pushes for level accuracy tasks and executing slow, quasi-static pushes for high accuracy tasks. 

We conduct experiments in simulation and on a real robotic system. We consider the tasks of pushing an object to a goal region and grasping an object in clutter. Given an environment for planning, we create two instantiations:
\vspace{-2mm}
\begin{itemize}
    \item[\textbf{$\square$}] \textbf{Planning environment:} The robot generates plans in the simulated planning environment. The trajectory optimizer (Alg.~\ref{alg:sto}) uses deterministic physics during planning and our online MDP solver uses stochastic physics to evaluate actions. 

    \item[$\square$] \textbf{Execution environment: }
Here, the robot executes actions and observes the state evolution. It is the physical world for real robot experiments but it is simulated for simulation experiments. The execution environment is stochastic. 
\end{itemize}

For the planning environment and the execution environment when it is simulated, we use
a physics engine, Mujoco\cite{mujoco}, 
to model the deterministic state transition function $f$ in Eq.~\ref{eq:dynamics}. We model stochasticity in the physics engine by adding Gaussian noise on the velocities $\{\dot{\vec{q}}^{R}, \dot{\vec{q}}^{1}, \dots, \dot{\vec{q}}^{D}\}$ of the robot and objects at every simulation time step:
\vspace{-2mm}
\begin{align}
\label{eq:simulation_noise}
\{\dot{\widetilde{\vec{q}}}^{R}, \dot{\widetilde{\vec{q}}}^{1}, \dots, \dot{\widetilde{\vec{q}}}^{D}\} =\{\dot{\vec{q}}^{R}, \dot{\vec{q}}^{1}, \dots, \dot{\vec{q}}^{D}\} + \vec{\mu}, \hspace{0.5cm} \vec{\mu} \sim \mathcal{N}(0,\beta(\vec{u_{t}}))
\end{align}

\noindent where $\mathcal{N}$ is the Gaussian distribution and $\beta$ is an action dependent variance function. We create a linear model for the variance function as: 
\vspace{-2mm}
\begin{align}
\beta(\vec{u}_{t}) = b ||\vec{u}_{t}||
\end{align}

\noindent Where $b$ is a constant. In our simulation experiments, \textit{uncertainty level} refers to the degree of stochasticity dictated by the 
slope $b$ of the variance function $\beta$ used to generate the Gaussian noise $\mu$ injected at every simulation time step in Eq.~\ref{eq:simulation_noise}. 
\begin{figure}[t]
	\begin{subfigure}[b]{0.49\textwidth}
		\includegraphics[height=1.7in, width=2.3in, angle=0]{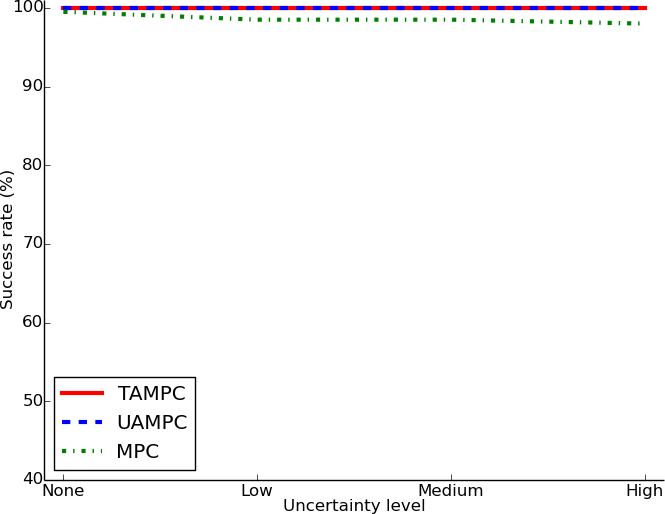}
		\caption{Success rates for low accuracy tasks}
		\label{fig:success_rate_low_accuracy}
	\end{subfigure}
	\vspace{5mm}
	\begin{subfigure}[b]{0.49\textwidth}
		\includegraphics[height=1.7in, width=2.3in, angle=0]{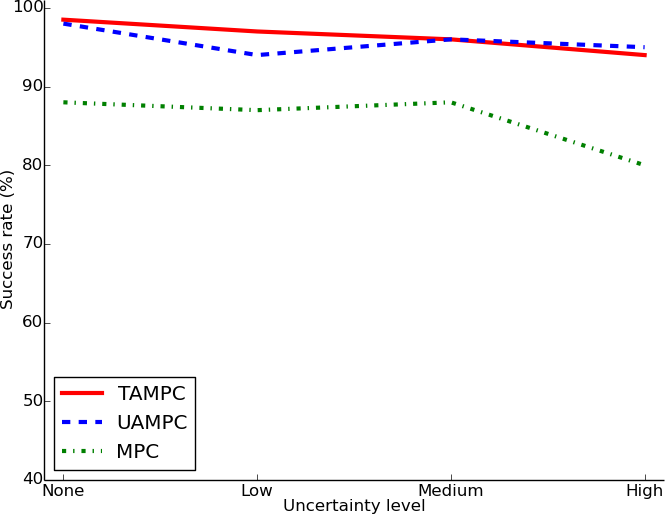}
		\caption{Success rates for high accuracy tasks}
		\label{fig:success_rate_high_accuracy}
	\end{subfigure}
	
	\begin{subfigure}[b]{0.49\textwidth}
		\includegraphics[height=1.7in, width=2.3in, angle=0]{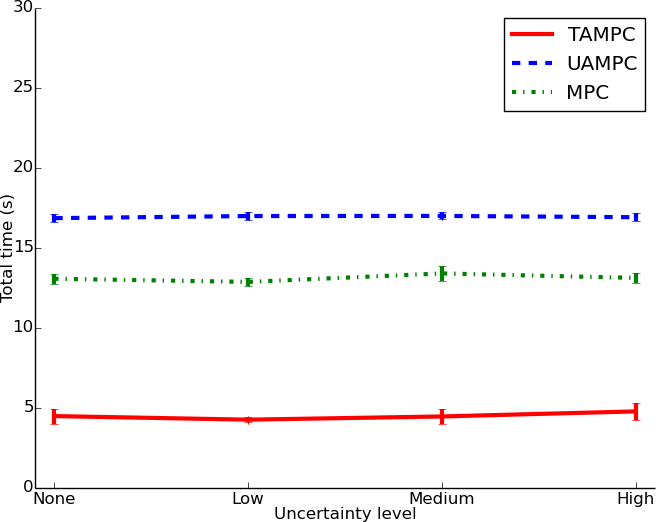}
		\caption{Total elapsed time for low accuracy tasks}
		\label{fig:total_time_low}
	\end{subfigure}
	\begin{subfigure}[b]{0.49\textwidth}
		\includegraphics[height=1.7in, width=2.3in, angle=0]{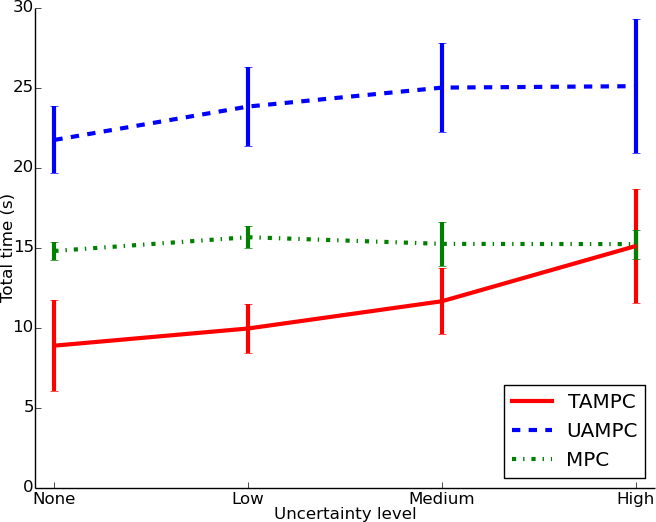}
		\caption{Total elapsed time for high accuracy tasks}
		\label{fig:total_time_high}
	\end{subfigure}
	\caption{Success rate and total elapsed time versus uncertainty level for low and high accuracy tasks. }
	  \end{figure}
	 \vspace{-3mm}
\subsection{Push planning simulation experiments}
\vspace{-2mm}
\label{sec:simulation}
\begin{figure}[t]
\centering 
  \begin{subfigure}[b]{0.245\textwidth}
    \includegraphics[height=1.18in,width=.6in, angle=90]{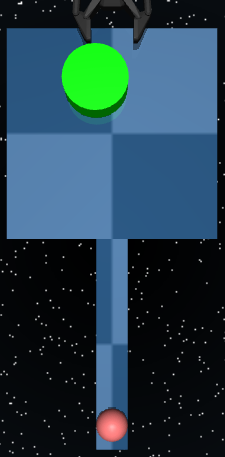}
    \caption*{Initial scene}
    %\label{fig:1}
  \end{subfigure}
  \hspace{-1.5mm}
  \begin{subfigure}[b]{0.245\textwidth}
    \includegraphics[height=1.18in,width=.6in, angle=90]{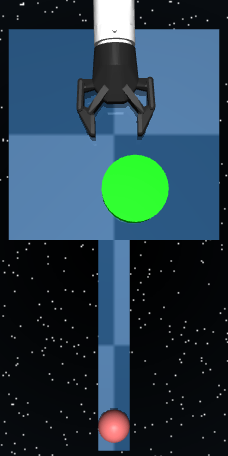}
    \caption*{After \textbf{1} action}
    %\label{fig:2}
  \end{subfigure}
  %\hspace{-7mm}
  \hspace{-1.5mm}
  \begin{subfigure}[b]{0.245\textwidth}
    \includegraphics[height=1.18in,width=.6in, angle=90]{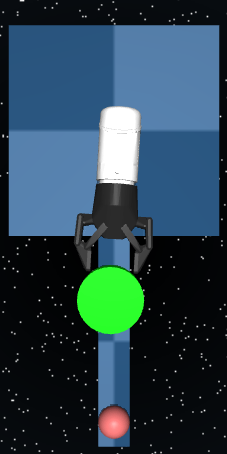}
    \caption*{\hspace{-5mm} After 8 actions}
    %\label{fig:3}
  \end{subfigure}
  %\hspace{-7mm}
  \hspace{-1.5mm}
  \begin{subfigure}[b]{0.245\textwidth}
    \includegraphics[height=1.18in,width=.6in, angle=90]{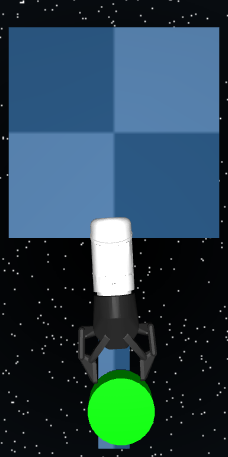}
    \caption*{\hspace{-5mm} Goal in 15 actions}
  \end{subfigure} 

    \begin{subfigure}[b]{0.245\textwidth}
    \includegraphics[height=.8in,width=1.15in]{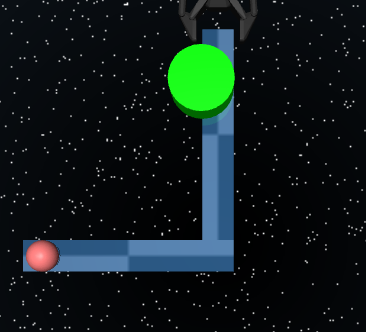}
    \caption*{Initial scene}
    \end{subfigure}
    \hspace{-2.2mm}
      \begin{subfigure}[b]{0.245\textwidth}
    \includegraphics[height=0.8in,width=1.15in]{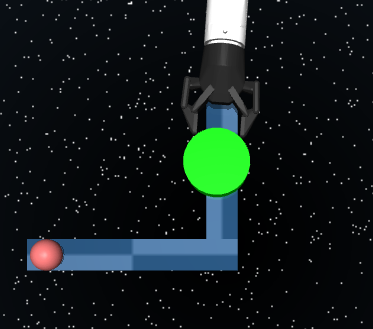}
    \caption*{After 10 actions}
  \end{subfigure} 
  \hspace{-2.2mm}
    \begin{subfigure}[b]{0.245\textwidth}
    \includegraphics[height=0.8in,width=1.15in]{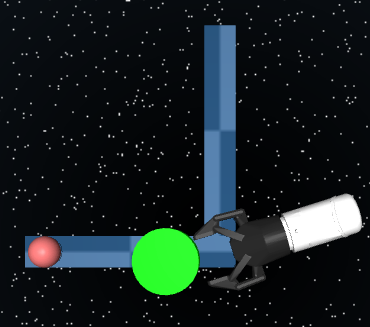}
    \caption*{After 40 actions}
  \end{subfigure} 
  \hspace{-2.2mm}
    \begin{subfigure}[b]{0.245\textwidth}
    \includegraphics[height=0.8in,width=1.15in]{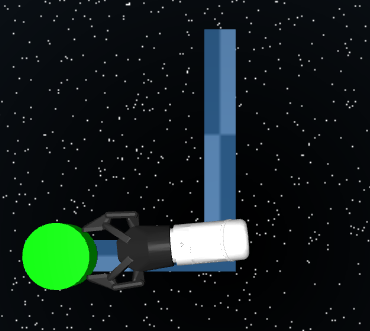}
    \caption*{Goal in\textbf{ 45 }actions}
  \end{subfigure} 
  \caption{Push planning in a changing environment (top) using a single fast push initially and then slow pushes later on due to the narrow strip. For the L-shaped environment (bottom), the robot executes many actions to successfully navigate the edge. 
  }
  \label{fig:change}
  %\vspace{-5mm}
   \end{figure}  

We present a high accuracy task in Fig.~\ref{fig:fig1} (top). It is made up of a thin strip and a small goal region. We also define a low accuracy task in Fig.~\ref{fig:fig1} (bottom) which is a much larger table with a wider goal region. We create 200 such planning environments for each of the high and low accuracy tasks. For each environment:  
\begin{enumerate}
\item We randomly select the shape (box or cylinder) of the pushed object. 
\item For each object, we randomly\footnote{
The uniform range used for each parameter is given here. 
Box x-y extents: $\left[0.05m,0.075m\right]$; 
box height: $\left[0.036m,0.05m\right]$;
cylinder radius:$\left[0.04m,0.07m\right]$;
cylinder height:$\left[0.04m,0.05m\right]$;
mass:$\left[0.2kg,0.8kg\right]$;
coef. fric.:$\left[0.2,0.6\right]$.
}  select
shape dimensions (radius and height for the cylinder, extents for the
boxes), mass, and coefficient of friction.
\item We randomly\footnote{ The random position for the pushed object is sampled from a Gaussian with a mean at the lower end of the table ($0.1m$ from the edge of a $0.6m$ long table along the center axis and a variance of $0.01 m$.)} select a position on the working surface for the pushed object. 

\end{enumerate}

We create four uncertainty levels: no uncertainty, low uncertainty, medium uncertainty and high uncertainty. For the no uncertainty case, no extra noise was added to the physics engine. For low, medium and high levels of uncertainty, $b=\{0.05,0.075,0.1\}$ respectively.  We test the different planning and control approaches and specify a timeout of 3 minutes including all planning, re-planning and execution.

%\vspace{3mm}

\noindent{\textbf{Success rates:}}
We declare success when the robot is able to push an object to the target region without dropping it off the edge of the table within the specified time limit. We plot the results in Fig.~\ref{fig:success_rate_low_accuracy} and ~Fig.~\ref{fig:success_rate_high_accuracy}.  For the low accuracy level push planning task (Fig.~\ref{fig:success_rate_low_accuracy}), $TAMPC$ and $UAMPC$ were able to maintain a 100 \% success rate while $MPC$ showed a slight decrease in success rates as uncertainty grew. For the high accuracy pushing task (Fig.~\ref{fig:success_rate_high_accuracy}), $TAMPC$ and $UAMPC$ were also able to maintain a good average success rate. $MPC$ on the other hand maintains a poor success rate. The major reason for this is uncertainty. 
%
\iffalse 
Recall that $MPC$ involves planning a set of actions to be applied on the robot sequentially for a fixed time horizon, the first action is applied and the process is repeated. The policy-lag (re-planning time) is high in this domain, typically requiring the robot to execute an action, stop and wait until the next action becomes available. If an action injects acceleration on a pushed object, it will continue moving after the robot has stopped to re-plan. Then, undesired events such as objects falling off the table can occur solely due to the policy-lag with $MPC$. We address this problem in $TAMPC$ (Sec.~\ref{sec:mdp}) during planning by waiting for objects to come to rest before computing the cost of an action. 
\fi 

\noindent{\textbf{Total time:}}
The total time in our experiments includes all planning and execution time. Fig~\ref{fig:total_time_low} and Fig~\ref{fig:total_time_high} show the average of 200 scenes with 95 \% confidence interval of the mean.  For the low accuracy level task, our $TAMPC$ planner is able to achieve the goal in under $5 s$ (Fig~\ref{fig:total_time_low}),  while $UAMPC$ and $MPC$ took  significantly more time to complete the task. This clearly shows that our method is able to generate successful fast actions while maintaining a high success rate. For the high accuracy level task (Fig~\ref{fig:total_time_high}), our planner is able to generate as many small actions as needed as the uncertainty grew. Hence it was able to maintain a high success rate and still complete the task within a very small amount of time in comparison with the baseline approach. 
\noindent Furthermore, we also test the adaptive behavior of our approach for the environments in Fig.~\ref{fig:change}. In the changing environment (Fig.~\ref{fig:change}, top), the robot begins with a fast push due to a large initial area. Thereafter, it naturally switches to slow pushes on the thin strip to complete the task. For pushing in the L-shaped environment (Fig.~\ref{fig:change}, bottom), the robot generally pushes slow. However, it spends a lot of time to navigate the corner.     
\vspace{-2mm}
\subsection{Grasping in clutter simulation experiments}
\begin{figure}[t]
\centering 
  \begin{subfigure}[b]{0.245\textwidth}
    \includegraphics[scale=0.31, angle=0]{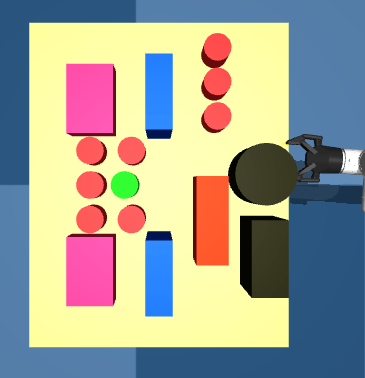}
    \caption*{Initial scene}
  \end{subfigure}
  \hspace{-1.9mm}
  \begin{subfigure}[b]{0.245\textwidth}
    \includegraphics[scale=0.308, angle=0]{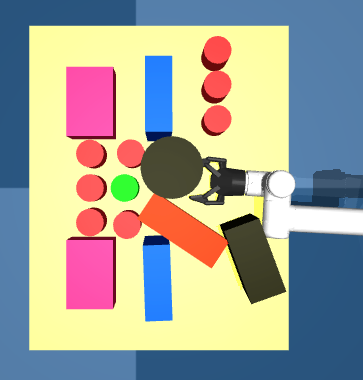}
    \caption*{After \textbf{3} actions}
  \end{subfigure}
  \hspace{-1.9mm}
  \begin{subfigure}[b]{0.245\textwidth}
    \includegraphics[scale=0.313, angle=0]{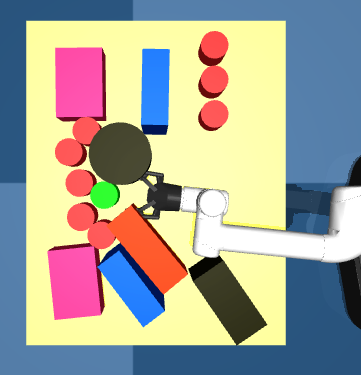}
    \caption*{\hspace{-5mm} After 7 actions}
  \end{subfigure}
  \hspace{-1.9mm}
  \begin{subfigure}[b]{0.245\textwidth}
    \includegraphics[scale=0.318, angle=0]{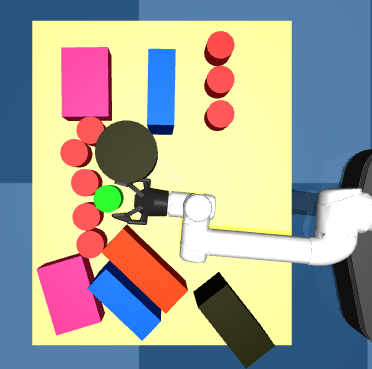}
    \caption*{\hspace{-5mm} Goal in 11 actions}
  \end{subfigure} 
  \vspace{-2mm}
  \caption{Grasping in clutter: The robot uses fast actions initially but chooses slower actions as it gets closer to the goal object near the edge of the table.}
  \label{fig:clutter_manipulation}
\end{figure}

\vspace{-1mm}
We conducted simulation experiments for grasping in clutter in scenes similar to Fig.~\ref{fig:clutter_manipulation}. Our scenes are randomly generated containing boxes and cylinders. In addition, our robot now has four control inputs (including the gripper). We tested our task adaptive planner in clutter to observe how the planner adapts given different environment configurations. We see that the robot manipulates clutter and is able to grasp the target object. An example scene is shown in Fig.~\ref{fig:clutter_manipulation} where the aim is to grasp the target object in green without pushing any other objects off the edge of the table. The robot initially begins with fast actions to push obstacles out of the way. However, as the robot gets closer to the target object, it chooses slower actions due to a higher probability of task failure in that region. 

\vspace{-3mm}

\subsection{Real robot experiments}
\begin{figure}[t]
\centering 
	\begin{subfigure}[b]{0.245\textwidth}
		\includegraphics[height=1.0in,width=1.15in]{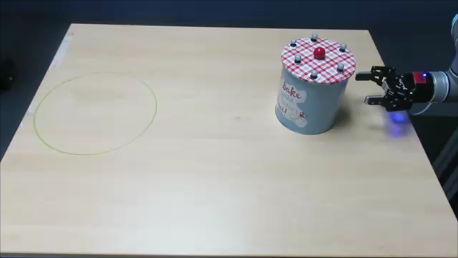}
   \begin{picture}(0,0)
      \put(-95.0,0.0){\rotatebox{90}{Low Accuracy Task}}
      \put(-75,50){  $\swarrow$ }
      \put(-75,58){\textbf{Goal}}
    \end{picture}
		\caption*{Initial scene}
		%\label{fig:low_task_accuracy}
	\end{subfigure}
	\hspace{-3mm}
	\begin{subfigure}[b]{0.245\textwidth}
		\includegraphics[height=1.0in,width=1.15in]{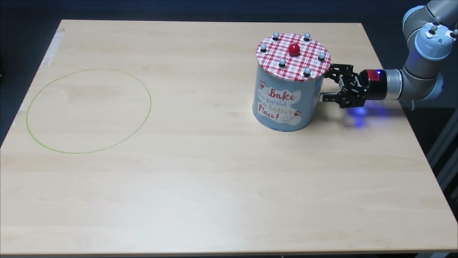}
		\caption*{After 5 actions}
		%\label{fig:2}
	\end{subfigure}
	\hspace{-3mm}
	\begin{subfigure}[b]{0.245\textwidth}
		\includegraphics[height=1.0in,width=1.15in]{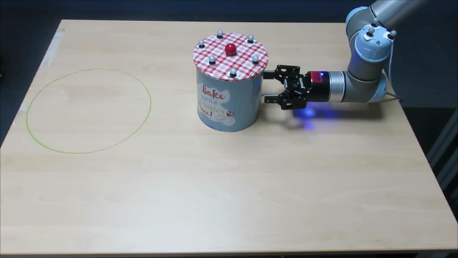}
		\caption*{After 12 actions}
		%\label{fig:3}
	\end{subfigure}
	\hspace{-3mm}
	\begin{subfigure}[b]{0.245\textwidth}
		\includegraphics[height=1.0in,width=1.15in]{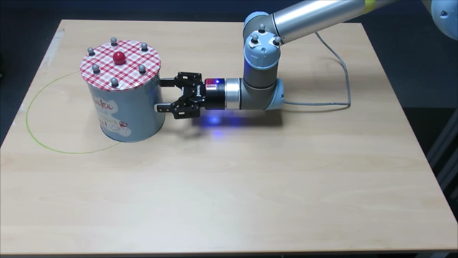}
		\caption*{Goal in 20 actions}
		%\label{fig:4}
	\end{subfigure} 
	\vspace{1mm}
	
	\begin{subfigure}[b]{0.245\textwidth}
		\includegraphics[height=1.0in,width=1.15in]{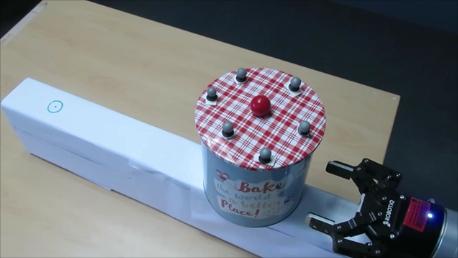}
	\begin{picture}(0,0)
      \put(-95.0,0.0){\rotatebox{90}{High Accuracy Task}}
       \put(-70,45){  $\swarrow$ }
      \put(-72,55){\textbf{Goal}}
    \end{picture}
		\caption*{Initial scene}
		%\label{fig:high_task_accuracy}
	\end{subfigure}
	\hspace{-3mm}
	\begin{subfigure}[b]{0.245\textwidth}
		\includegraphics[height=1.0in,width=1.15in]{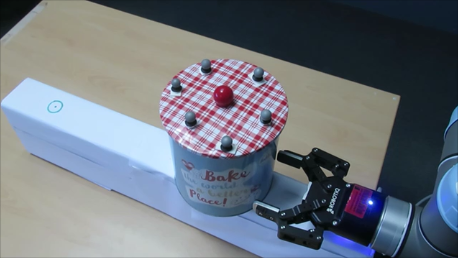}
		\caption*{After 5 actions}
		%\label{fig:2}
	\end{subfigure}
	\hspace{-3mm}
	\begin{subfigure}[b]{0.245\textwidth}
		\includegraphics[height=1.0in,width=1.15in]{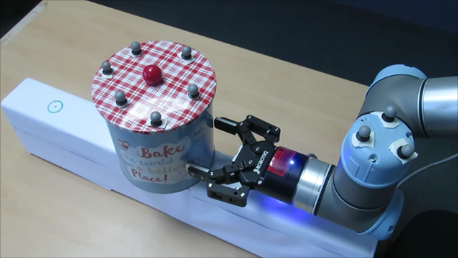}
		\caption*{After 10 actions}
		%\label{fig:3}
	\end{subfigure}
	\hspace{-3mm}
	\begin{subfigure}[b]{0.245\textwidth}
		\includegraphics[height=1.0in,width=1.15in]{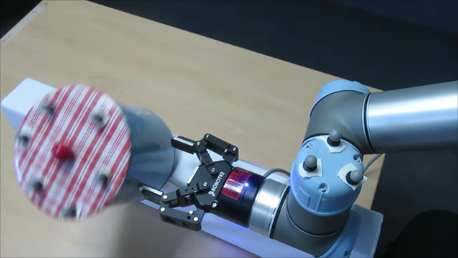}
		\caption*{Failed in 13 actions }
		%\label{fig:smpc_low}
	\end{subfigure} 
	\vspace{-2mm}
	\caption{$MPC$ using a large number of actions to complete a low accuracy level task (top), and causing the pushed object to fall off for a high accuracy level task (bottom). }
	\label{fig:standard_mpc}
\end{figure}
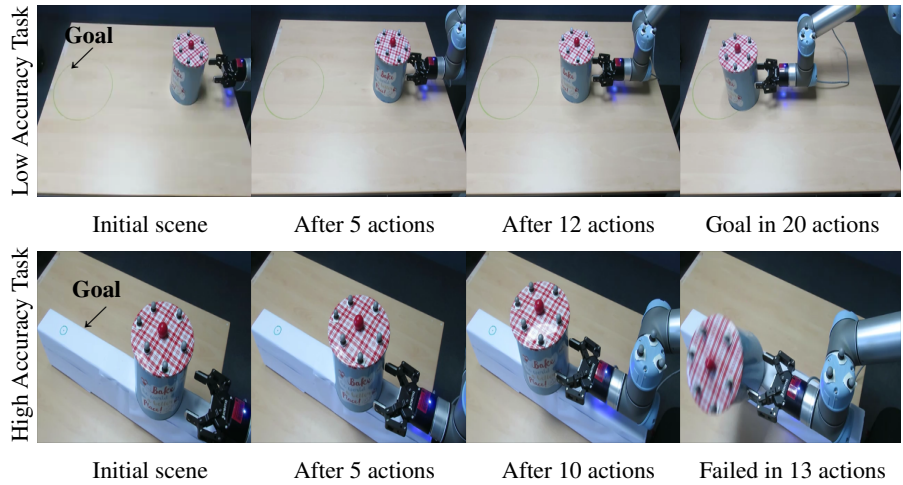

\vspace{-2mm}
\label{sec:realrobot}
%\vspace{-1.1mm}
In the real robot experiments we use a UR5 robot with a Robotiq $2$ finger gripper. We restrict the motion of the gripper to a plane parallel to the working surface such that we have a planar robot. We use OpenRave\citep{Diankov} to find kinematics solutions at every time step.
For the push planning experiments, the gripper is completely open such that the robot receives three control inputs $\vec{u}_t=(\dot{\theta}_x,\dot{\theta}_y, \dot{\theta}_{rotation} )$ at every time step. We use a medium uncertainty level to model the real world stochasticity. We place markers on the pushed object and track its full pose with a motion capture system (OptiTrack). We manually replicated three execution worlds for each task accuracy level from the randomly generated environments we created during push planning simulation experiments. We tested our planners in these environments. We show snapshots from our real robot experiments. In Fig.~\ref{fig:standard_mpc} (top), we have a low task accuracy environment where the standard MPC approach is successful after 20 actions. However, by using a single dynamic push in Fig.~\ref{fig:fig1} (bottom), our task-adaptive control approach is able to complete the push planning task in under 2 seconds. 

Moreover, for the high task accuracy problem, $MPC$ was unable to push the target object to the desired goal location (Fig.~\ref{fig:standard_mpc} (bottom)). It executes actions without reasoning about uncertainty and pushed the goal object off the edge. Our task-adaptive controller was able to consider uncertainty while generating small pushes (Fig.~\ref{fig:fig1} (top)) to complete the task. These results can be found in the accompanying video at \url{https://youtu.be/8rz_f_V0WJA}. 

% push planning snapshots 
\vspace{-3mm}
\section{RELATED WORK}
\vspace{-4mm}
Robots are becoming increasingly capable of performing complex manipulation tasks \cite{Zhu-RSS-18} by leveraging the environment's physics \cite{Toussaint-RSS-18}. A traditional approach to complete manipulation tasks is motion planning \cite{ilqr, PRM}, followed by open-loop execution. However, these plans are likely to fail due to uncertainty. Others \cite{Sieverling_uncertainty, zhou_mason, convergent_planning, cc_RRT} have proposed using conservative open-loop motion plans that are robust to uncertainty. Recently, MPC algorithms for manipulation \cite{mppi_push, pusher_slider, Dollar_ICRA17, real_time_Agboh_Humanoids18} have been developed. They use feedback and can handle bounded uncertainty \cite{MAYNE_MPC}. We take a similar closed-loop approach to non-prehensile manipulation in this paper. 

There are recent works that develop learning-based uncertainty-aware controllers in robotics mainly for navigation and collision avoidance. \citet{charles_RSS17} proposed safe visual navigation with deep learning. Given that current deep learning methods produce unsafe predictions when faced with out-of-distribution inputs, they detect these novel inputs using the reconstruction loss of an autoencoder and switch to a rule-based safe policy. Similarly, \citet{choi_ICRA18} switch between a learned policy and a rule-based policy for autonomous driving. However, they estimate uncertainty using a single mixture density network without Monte Carlo sampling. We focus on non-prehensile manipulation where designing a rule-based policy for each new task is not feasible. 

Furthermore, \citet{greg_collision17} proposed an uncertainty-aware controller for autonomous navigation with collision avoidance. First, they estimate uncertainty from multiple bootstrapped neural networks using dropout. Thereafter, they consider a very large number of fixed action sequences at a given current state. They evaluate these action sequences under an uncertainty-dependent cost function within an MPC framework. The resulting policy chooses low speed actions in unfamiliar environments and naturally chooses high speed actions in familiar environments. While this approach relieves the burden of designing a rule-based policy, it requires the evaluation of a large number of \textit{a priori} fixed action sequences. 

% \iffalse
% In non-prehensile manipulation, especially in cluttered environments, the cost of physics predictions is high. Hence, given that our goal is closed-loop pushing in real-time, it is not practical to consider a large number of action sequences at a given state. In this work, we do not define a fixed action set \textit{a priori}. Instead, for a given current state we sample promising parts of the action space using a trajectory optimizer. Thus, we consider a small set of promising actions including both fast and slow actions. In this way, we are able to achieve closed-loop adaptive pushing under uncertainty in real-time: pushing fast when the task permits inaccuracy and naturally pushing slow for more difficult tasks. 
% \fi

Our approach of adaptively sampling the action space given a current state is similar to prior work \cite{x_armed_bandits, littman} which propose sampling-based planning in continuous action MDPs. We use a trajectory optimizer to guide our action sampling process.

%where the action space of an MDP is allowed to be a generic topological space. 
%
%{\let\clearpage\relax  \section{DISCUSSION AND FUTURE WORK}}
\vspace{-2mm}
\section{DISCUSSION AND FUTURE WORK}
\vspace{-3mm}
We presented a closed-loop planning and control algorithm capable of adapting to the accuracy requirements of a task by generating both fast and slow actions.This is an exciting first step toward realizing task-adaptive manipulation planners. In this work, we use a stochastic trajectory optimizer that outputs locally optimal control sequences. Thus, the resulting policy can get stuck in a local optima. Moreover, the trajectory optimizer may not return a good control sequence that reaches the goal for a given task if some design parameters (e.g. $n_{max}$) are chosen poorly. 
Furthermore, our uncertainty model is simple and is only an approximation of the real world stochastic phenomena. We will work toward finding uncertainty models that better describe the real world physics stochasticity especially for manipulation in clutter.
Finally, we will investigate the generalization of our task-adaptive approach to other manipulation primitives. 
\vspace{-2mm}
\section{ACKNOWLEDGEMENTS}
\vspace{-3mm}
This project has received funding from the European Union’s Horizon 2020 research and innovation programme under the Marie Sklodowska Curie grant agreement No. 746143, and from the UK Engineering and Physical Sciences Research Council under grants EP/P019560/1 and EP/R031193/1.

\vspace{-1mm}
\vspace{5mm}
\bibliographystyle{plainnat}
\hspace{-100mm}
\vspace{-5mm}
{\let\clearpage\relax \bibliography{bibliography_file} }

\begin{thebibliography}{33}
\providecommand{\natexlab}[1]{#1}
\providecommand{\url}[1]{\texttt{#1}}
\expandafter\ifx\csname urlstyle\endcsname\relax
  \providecommand{\doi}[1]{doi: #1}\else
  \providecommand{\doi}{doi: \begingroup \urlstyle{rm}\Url}\fi

\bibitem[Agboh and Dogar(2018)]{real_time_Agboh_Humanoids18}
Wisdom~C. Agboh and Mehmet~R. Dogar.
\newblock Real-time online re-planning for grasping under clutter and
  uncertainty.
\newblock In \emph{IEEE-RAS Humanoids}, 2018.

\bibitem[Arruda et~al.(2017)Arruda, Mathew, Kopicki, Mistry, Azad, and
  Wyatt]{mppi_push}
E.~Arruda, M.~J. Mathew, M.~Kopicki, M.~Mistry, M.~Azad, and J.~L. Wyatt.
\newblock Uncertainty averse pushing with model predictive path integral
  control.
\newblock In \emph{Humanoids}, 2017.

\bibitem[Bubeck et~al.(2009)Bubeck, Stoltz, Szepesv\'{a}ri, and
  Munos]{x_armed_bandits}
S\'{e}bastien Bubeck, Gilles Stoltz, Csaba Szepesv\'{a}ri, and R\'{e}mi Munos.
\newblock Online optimization in x-armed bandits.
\newblock In \emph{NIPS}. 2009.

\bibitem[Calli and Dollar(2017)]{Dollar_ICRA17}
B.~Calli and A.~M. Dollar.
\newblock Vision-based model predictive control for within-hand precision
  manipulation with underactuated grippers.
\newblock In \emph{ICRA}, 2017.

\bibitem[Choi et~al.(2018)Choi, Lee, Lim, and Oh]{choi_ICRA18}
S.~Choi, K.~Lee, S.~Lim, and S.~Oh.
\newblock Uncertainty-aware learning from demonstration using mixture density
  networks with sampling-free variance modeling.
\newblock In \emph{ICRA}, 2018.

\bibitem[Diankov et~al.(2008)Diankov, Srinivasa, Ferguson, and
  Kuffner]{Diankov}
R.~Diankov, S.~S. Srinivasa, D.~Ferguson, and J.~Kuffner.
\newblock Manipulation planning with caging grasps.
\newblock In \emph{Humanoids}, 2008.

\bibitem[Dogar et~al.(2012)Dogar, Hsiao, Ciocarlie, and
  Srinivasa]{dogar_clutter}
M.~R Dogar, K.~Hsiao, M.~Ciocarlie, and S.~Srinivasa.
\newblock Physics-based grasp planning through clutter.
\newblock In \emph{Robotics: Science and Systems}, 2012.

\bibitem[Fitts(1954)]{fitts_law}
Paul~M. Fitts.
\newblock The information capacity of the human motor system in controlling the
  amplitude of movement.
\newblock \emph{Experimental Psychology}, 1954.

\bibitem[Hogan and Rodriguez(2016)]{pusher_slider}
F.~R. Hogan and A.~Rodriguez.
\newblock Feedback control of the pusher-slider system: {A} story of hybrid and
  underactuated contact dynamics.
\newblock \emph{WAFR}, 2016.

\bibitem[Howe and Cutkosky(1996)]{howe1996practical}
Robert~D Howe and Mark~R Cutkosky.
\newblock Practical force-motion models for sliding manipulation.
\newblock \emph{IJRR}, 15\penalty0 (6):\penalty0 557--572, 1996.

\bibitem[Johnson et~al.(2016)Johnson, King, and Srinivasa]{convergent_planning}
A.~M. Johnson, J.~King, and S.~Srinivasa.
\newblock Convergent planning.
\newblock \emph{IEEE RA-L}, 2016.

\bibitem[Kahn et~al.(2017)Kahn, Villaflor, Pong, Abbeel, and
  Levine]{greg_collision17}
Gregory Kahn, Adam Villaflor, Vitchyr Pong, Pieter Abbeel, and Sergey Levine.
\newblock Uncertainty-aware reinforcement learning for collision avoidance.
\newblock \emph{CoRR}, 2017.

\bibitem[Kalakrishnan et~al.(2011)Kalakrishnan, Chitta, Theodorou, Pastor, and
  Schaal]{kalakrishnan2011stomp}
M.~Kalakrishnan, S.~Chitta, E.~Theodorou, P.~Pastor, and S.~Schaal.
\newblock Stomp: Stochastic trajectory optimization for motion planning.
\newblock In \emph{ICRA}, 2011.

\bibitem[Kavraki and Latombe(1994)]{PRM}
L.~Kavraki and J.~C. Latombe.
\newblock Randomized preprocessing of configuration for fast path planning.
\newblock In \emph{ICRA}, 1994.

\bibitem[Kearns et~al.(1999)Kearns, Mansour, and Ng]{kearns}
Michael~J. Kearns, Yishay Mansour, and Andrew~Y. Ng.
\newblock A sparse sampling algorithm for near-optimal planning in large markov
  decision processes.
\newblock In \emph{IJCAI}, 1999.

\bibitem[King et~al.(2015)King, Haustein, Srinivasa, and
  Asfour]{king2015nonprehensile}
J.~E. King, J.~A. Haustein, S.~Srinivasa, and T.~Asfour.
\newblock Nonprehensile whole arm rearrangement planning on physics manifolds.
\newblock In \emph{ICRA}, 2015.

\bibitem[Kitaev et~al.(2015)Kitaev, Mordatch, Patil, and Abbeel]{kitaev_abbeel}
N.~Kitaev, I.~Mordatch, S.~Patil, and P.~Abbeel.
\newblock Physics-based trajectory optimization for grasping in cluttered
  environments.
\newblock In \emph{ICRA}, 2015.

\bibitem[Li and Todorov(2004)]{ilqr}
Weiwei Li and Emanuel Todorov.
\newblock Iterative linear quadratic regulator design for nonlinear biological
  movement systems.
\newblock In \emph{ICINCO}, 2004.

\bibitem[Luders et~al.(2010)Luders, Kothari, and How]{cc_RRT}
B.~Luders, M.~Kothari, and J.P. How.
\newblock Chance constrained rrt for probabilistic robustness to environmental
  uncertainty.
\newblock In \emph{AIAA Guidance, Navigation, and Control Conference}, 2010.

\bibitem[Mansley et~al.(2011)Mansley, Weinstein, and Littman]{littman}
C.~Mansley, A.~Weinstein, and M.~L. Littman.
\newblock Sample-based planning for continuous action markov decision
  processes.
\newblock In \emph{ICAPS}, 2011.

\bibitem[Mason(1986)]{mason_pushing}
Matthew~T. Mason.
\newblock Mechanics and planning of manipulator pushing operations.
\newblock \emph{The International Journal of Robotics Research}, 5\penalty0
  (3):\penalty0 53--71, 1986.

\bibitem[Mayne et~al.(2000)Mayne, Rawlings, Rao, and Scokaert]{MAYNE_MPC}
D.Q. Mayne, J.B. Rawlings, C.V. Rao, and P.O.M. Scokaert.
\newblock Constrained model predictive control: Stability and optimality.
\newblock \emph{Automatica}, 36\penalty0 (6):\penalty0 789 -- 814, 2000.

\bibitem[Muhayyuddin et~al.(2018)Muhayyuddin, Moll, Kavraki, and
  Rosell]{randomized_clutter_uncertainty}
Muhayyuddin, M.~Moll, L.~Kavraki, and J.~Rosell.
\newblock Randomized physics-based motion planning for grasping in cluttered
  and uncertain environments.
\newblock \emph{IEEE RA-L}, 3\penalty0 (2):\penalty0 712--719, 2018.

\bibitem[P{\'e}ret and Garcia(2004)]{peret2004line}
Laurent P{\'e}ret and Fr{\'e}d{\'e}rick Garcia.
\newblock On-line search for solving markov decision processes via heuristic
  sampling.
\newblock In \emph{ECAI}. IOS Press, 2004.

\bibitem[Richter and Roy(2017)]{charles_RSS17}
Charles Richter and Nicholas Roy.
\newblock Safe visual navigation via deep learning and novelty detection.
\newblock In \emph{RSS}, 2017.

\bibitem[Ruiz-Ugalde et~al.(2011)Ruiz-Ugalde, Cheng, and
  Beetz]{effect_aware_push}
F.~Ruiz-Ugalde, G.~Cheng, and M.~Beetz.
\newblock Fast adaptation for effect-aware pushing.
\newblock In \emph{Humanoids}, 2011.

\bibitem[Sieverling et~al.(2017)Sieverling, Eppner, Wolff, and
  Brock]{Sieverling_uncertainty}
A.~Sieverling, C.~Eppner, F.~Wolff, and O.~Brock.
\newblock Interleaving motion in contact and in free space for planning under
  uncertainty.
\newblock In \emph{IROS}, 2017.

\bibitem[Todorov et~al.(2012)Todorov, Erez, and Tassa]{mujoco}
E.~Todorov, T.~Erez, and Y.~Tassa.
\newblock Mujoco: A physics engine for model-based control.
\newblock In \emph{IROS}, 2012.

\bibitem[Toussaint et~al.(2018)Toussaint, Allen, Smith, and
  Tenenbaum]{Toussaint-RSS-18}
M.~Toussaint, K.~Allen, K.~Smith, and J.~Tenenbaum.
\newblock Differentiable physics and stable modes for tool-use and manipulation
  planning.
\newblock In \emph{RSS}, 2018.

\bibitem[Williams et~al.(2015)Williams, Aldrich, and Theodorou]{mppi}
G.~Williams, A.~Aldrich, and E.~Theodorou.
\newblock Model predictive path integral control using covariance variable
  importance sampling.
\newblock \emph{CoRR}, 2015.

\bibitem[Yu et~al.(2016)Yu, Bauza, Fazeli, and Rodriguez]{million_ways_to_push}
K.~T. Yu, M.~Bauza, N.~Fazeli, and A.~Rodriguez.
\newblock More than a million ways to be pushed. a high-fidelity experimental
  dataset of planar pushing.
\newblock In \emph{IROS}, 2016.

\bibitem[Zhou et~al.(2017)Zhou, Paolini, Johnson, Bagnell, and
  Mason]{zhou_mason}
J.~Zhou, R.~Paolini, A.~M. Johnson, J.~A. Bagnell, and M.~T. Mason.
\newblock A probabilistic planning framework for planar grasping under
  uncertainty.
\newblock \emph{IEEE RA-L}, 2\penalty0 (4):\penalty0 2111--2118, 2017.

\bibitem[Zhu et~al.(2018)Zhu, Wang, Merel, Rusu, Erez, Cabi, Tunyasuvunakool,
  Kramár, Hadsell, de~Freitas, and Heess]{Zhu-RSS-18}
Y.~Zhu, Z.~Wang, J.~Merel, A.~Rusu, T.~Erez, S.~Cabi, S.~Tunyasuvunakool,
  J.~Kramár, R.~Hadsell, Nando de~Freitas, and N.~Heess.
\newblock Reinforcement and imitation learning for diverse visuomotor skills.
\newblock In \emph{RSS}, 2018.

\end{thebibliography}
\vspace{-5mm}
\end{document}